\RequirePackage{amsmath}
\documentclass[runningheads]{llncs}

\usepackage[utf8]{inputenc}
\usepackage[T1]{fontenc}
\usepackage{amsfonts}
\usepackage{cite}

\usepackage{graphicx}
\usepackage{url}
\usepackage{hyperref}
\usepackage{color}

\begin{document}

\title{Symbolic Integration Algorithm Selection with Machine Learning: LSTMs vs Tree LSTMs}
%\titlerunning{Abbreviated paper title}
% If the paper title is too long for the running head, 
% you can set an abbreviated paper title here

\author{Rashid Barket\inst{1}\orcidID{0000-0002-9104-4281} \and 
Matthew England\inst{1}\orcidID{0000-0001-5729-3420} \and
J\"{u}rgen Gerhard\inst{2}}
\authorrunning{Barket et al.}
% First names are abbreviated in the running head.
% If there are more than two authors, 'et al.' is used.
%
\institute{
Coventry University, Coventry, United Kingdom\\
\email{\{barketr, matthew.england\}@coventry.ac.uk} \and 
Maplesoft, Waterloo, Ontario, Canada\\
\email{jgerhard@maplesoft.com}}

\maketitle              % typeset the header of the contribution
\begin{abstract}
Computer Algebra Systems (e.g. Maple) are used in research, education, and industrial settings. One of their key functionalities is symbolic integration, where there are many sub-algorithms to choose from that can affect the form of the output integral, and the runtime. Choosing the right sub-algorithm for a given problem is challenging: we hypothesise that Machine Learning can guide this sub-algorithm choice. 

A key consideration of our methodology is how to represent the mathematics to the ML model:  we hypothesise that a representation which encodes the tree structure of mathematical expressions would be well suited.  We trained both an LSTM and a TreeLSTM model for sub-algorithm prediction and compared them to Maple's existing approach. 

Our TreeLSTM performs much better than the LSTM, highlighting the benefit of using an informed representation of mathematical expressions. It is able to produce better outputs than Maple's current state-of-the-art meta-algorithm, giving a strong basis for further research.

\end{abstract}

\section{Introduction}
\label{sec:intro}

Machine Learning (ML), and specifically deep learning, has seen a surge of applications in many domains, but only recently have there been applications to computer algebra. One can take two possible approaches when using ML in this field: to directly make predictions to solve a problem, or to aid existing algorithms in their free choices to improve an objective function. 

% A Computer Algebra System (CAS) is known for its precise calculations. If a CAS were to make an incorrect calculation, it would be detrimental to both the user and the CAS itself. On the other hand, ML has a probabilistic nature and tends to predict what an answer to a problem may be. Thus, it seems as though ML and CASs would have little to no interaction. However, for many implemented CA algorithms in a CAS, there is usually a choice that can be made. When an algorithm has to make a choice, this is either done in a pre-set order or a deterministic heuristic. With the way these choices are made, computation time could be wasted or non-optimal answers may be produced.

\begin{figure}[ht]
    \centering
    \includegraphics[scale=0.2]{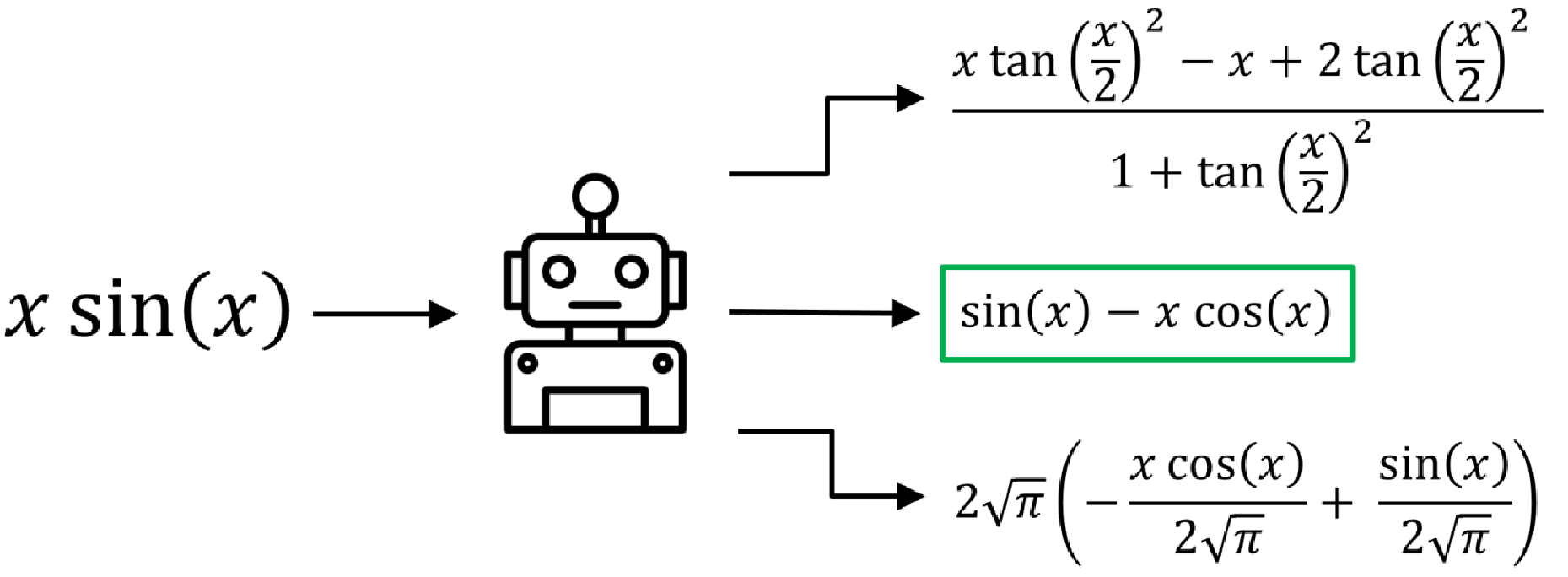}
    \caption{The output of $\int  x\sin(x)$ from three successful sub-algorithm. The optimal output is the shortest expression from the second sub-algorithm.}
    \label{fig:sub-algo-select}
\end{figure}

We focus on the Computer Algebra System (CAS) Maple, and its main symbolic integration algorithm, \texttt{int}, which is essentially a meta-algorithm to choose from 12 possible sub-algorithms Maple has for indefinite integration\footnote{\url{https://www.maplesoft.com/support/help/Maple/view.aspx?path=int}}. The names of each sub-algorithm are also available later in Figure \ref{fig:data_freq}. Each can produce very different, but mathematically equivalent answers, as in the example in Figure \ref{fig:sub-algo-select}. Some can also take much longer to execute than others. 

In this paper, we will train LSTM and TreeLSTM models to select the sub-algorithm that produces the optimal length answer for given problems. This is important to a Maple user who would prefer a simpler expression if available. Because the objective function is based on the output length, multiple sub-algorithms may be optimal. Thus, this is a multi-label classification problem.% where we can have that more than one answer is correct for a selection. 
%We show that the TreeLSTM model has an improvement over Maple's meta-algorithm even after training on a small dataset. 

We proceed with a brief literature review in Section \ref{sec:LR}, before outlining our machine learning methodology and data generation processes in Sections \ref{sec:ML} and \ref{sec:Data}.  We give our experimental results in Section \ref{sec:Results} and then conclude in Section \ref{sec:Conclusion}.  

%The rest of the paper is outlined as follows: We give a brief explanation of the key concepts from ML used in this model creation. This includes the Long-Short Term Memory (LSTM) model and its variations, multi-label classification \& classifier chains, and different metrics for evaluating the models. Then, a full description of the problem we are attempting to solve, improvement to the output of symbolic integration, is given. Here, we also talk about how to represent mathematical expressions for the ML model and how data is created. We then look at the results of our trained model and how it compares to Maple’s current meta-algorithm. Lastly, we talk about current limitations and future work.  

\section{Literature Review}
\label{sec:LR}

The problem of using ML to improve computer algebra algorithms has gained traction within the last decade. One of the first uses was for cylindrical algebraic decomposition algorithm where the choice of variable ordering can be key to tractability. Huang et al. made the first attempt in 2014 using a Support Vector Machine \cite{Huang2014}, with more recent work on this problem involving reinforcement learning and graph neural networks \cite{Jia2023} and Explainable AI techniques \cite{Pickering2024}. 

The most relevant work for our problem was by Lample \& Charton \cite{Lample2020} who trained a transformer to calculate integrals directly, learning from a large quantity of (integrand, integral) pairs.
%A dataset is created with three different data generators and the transformer is trained on a large quantity of (integrand, integral) pairs and the differential equations and their solutions. 
They could calculate some integrals that CASs could not, but there are critiques of their data generation and testing methods \cite{Piotrowski2019}. %Although we are not trying to predict integrals directly, the data generation methods and model architecture are still reelvant to our problem. 

% Another popular area being studied is machine learning for improving Buchberger's algorithm. Like Cylindrical Algebraic Decomposition, this algorithm also has a doubly exponential time in the worst case. The bottleneck in this algorithm stems from choosing a pair of polynomials from a set of possible choices. Peifer et al. \cite{Peifer2020} were the first to introduce a reinforcement learning approach to selecting the pair of polynomials with better success than hand-crafted heuristics. There has also been research done in predicting a Groebner bases directly \cite{Jia2023} and in the number of polynomial additions done during Buchberger's algorithm \cite{Mojsilovic2023}. 

Simpson et al.~ gave an example of ML for computer algebra algorithm selection in  \cite{Simpson2016}: training ML to select from four algorithms to calculate the resultant of two polynomials, reducing CPU time in both Maple and Mathematica. 
%This is similar to what we aim but with two differences: We have 12 algorithms to choose from instead of 4, and we are forming a multi-label classification problem rather than a multi-class one.

% Expression simplification using ML has been explored in a few different formats. Shi et al. \cite{Shi2020_simplification} tackle how to simplify symbolic expressions without the need for human-made heuristics to be included in the dataset or model. They show that their model can discover more simplification rules than other methods that require human input. Another approach using reinforcement learning and transformers was explored by Dersey et al. \cite{Dersey2023_simplification} to decide which rules to use in simplifying polylogarithms (expressions with logarithms and dilogarithms). They can get both approaches to work successfully but had even more success with Transformers with a 91\% accuracy from their dataset. The authors also discuss how transformers have a bigger potential in the generative task of discovering new identities for simplifying polylogarithms. 

\section{Machine Learning}
\label{sec:ML}

When integrating with the 12 different sub-algorithms in Maple, each one can produce an answer or output failure. Our objective function, chosen in consultation with Maplesoft, is to minimise the length of the output. With this objective, multiple sub-algorithms may be optimal making this a multi-label classification problem. %See e.g. the scikit-multilearn library \cite{Szymanski2019_skmultilearn}. 

The baseline approach of binary relevance is used in this experiment. With $L$ is the set of possible labels, we train $|L|$ binary classifiers $C_1,...,C_{|L|}$. Each $C_j$ is responsible for predicting the binary classification for each $l_j \in L$. Each binary model $C_j$ is trained independently from the rest.  We will consider two ML models to tackle the problem introduced in the following sections.

This problem has not been tackled by ML before. To judge the quality of our models we will compare them to Maple's meta-algorithm implemented for \texttt{int}.

\subsection{LSTM}

Long-Short Term Memory (LSTM) networks were proposed in \cite{Horchreiter1997_LSTM}, and were the model of choice for many natural language tasks such as sentiment analysis and machine translation until transformers rose to popularity more recently. Like recurrent neural networks, LSTMs calculate an output $y_t$ based on the input at that time step, $x_t$, and the hidden state from the previous step, $h_{t-1}$. The key difference is the inclusion of a long-term memory which is maintained and updated over a long sequence of time. 
Three new gating mechanisms (the forget, input, and output gates) are introduced to maintain this long-term memory. 
As sub-algorithm selection for symbolic integration is not a task that has been explored before, there is no baseline to compare to. Thus, we will use LSTM as the baseline to compare against other models in this and future research.

\subsection{TreeLSTM}

A limitation of the LSTM network is that it can only process sequential information (e.g. text, audio, time-series data). We hypothesise that embedding data with a tree structure (how a CAS like Maple stores mathematical expressions) would be beneficial.

\begin{figure}[b]
    \centering
    \includegraphics[width=0.7\textwidth]{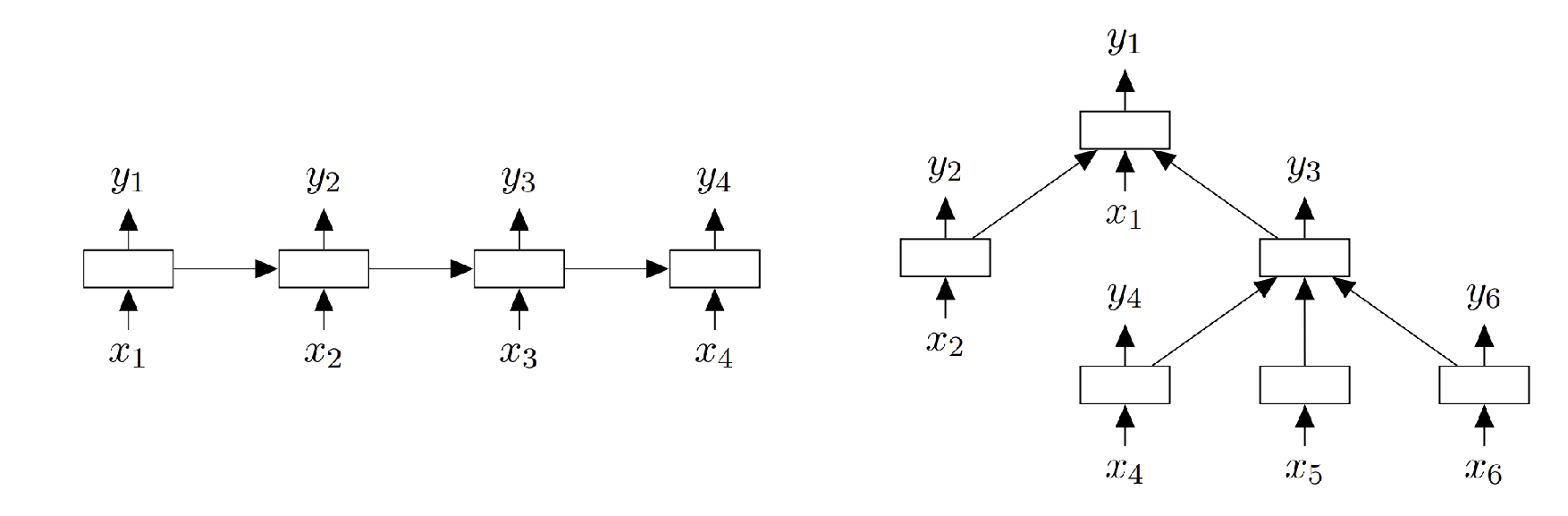}
    \caption{Visual representation of an LSTM (left) and TreeLSTM (right) \cite{Tai2015_TreeLSTM}. 
    %The LSTM creates its hidden state $h_t$ and output $y_t$ based on an input vector $x_t$ and hidden state from the previous cell $h_{t-1}$. The TreeLSTM creates its hidden state and output based on an input vector $x_t$ and arbitrarily many children $h_1,...,h_n$.
    }
    \label{fig:TreeLSTM}
\end{figure}

TreeLSTM is a variant of the LSTM network proposed by Tai et al. \cite{Tai2015_TreeLSTM} for tree-structured inputs. Like LSTM, it contains a long-term memory with a gating architecture. However, the hidden state and cell memory of step $t$ is dependent on arbitrarily many children instead of a single child from step $t-1$. There are forget gates for each child so the network can learn which information is important from which child to add to the hidden state and cell memory. The structure of the TreeLSTM is compared to LSTM in Figure \ref{fig:TreeLSTM}. 

%The overall goal is to be able to select a sub-algorithm that produces the shortest expression from a set of sub-algorithms available for symbolic integration. As previously mentioned, this is an experiment that has not been done for this task. While we can compare which model does the best, what we want to show is that a classifier does a better job at this task than actual CASs. To this end, we will use the predictions generated by the classifier and see how those predictions compare to the meta-algorithm implemented in the CAS Maple. This is a similar approach taken by Lample \& Charton \cite{Lample2020} where they compared the results of their integration and differential equation solving tasks to Maple and Mathematica. 

\section{Generating Data}
\label{sec:Data}

As with any ML problem we need a source of data, in our case integrable mathematical expressions.  We need a rich variety so that the model can generalise. 

\subsection{Existing Methods}
\label{sec:data/exisiting}

Lample \& Charton introduced three data generation methods in \cite{Lample2020}:
\begin{itemize}
    \item \textbf{FWD:} Integrate an expression $f$ through a CAS to get $F$ and add the pair ($f, F$) to the dataset.
    \item \textbf{BWD:} Differentiate $f$ to get $f'$ and add the pair ($f', f$) to the dataset.
    \item \textbf{IBP:} Given two expressions $f$ and $g$, calculate $f'$ and $g'$. If $\int f'g$ is known then the following holds (integration-by-parts):
    $\int fg' = fg - \int f'g$.
    Thus, we add the pair ($fg'$, $fg - \int f'g$) to the dataset. 
\end{itemize}
These offer a starting point but do not generate the rich variety we need.  A thorough discussion of the shortcomings of these methods was given in \cite{Barket2023_generation}.

% so we will discuss them at a high-level overview. For both the FWD and IBP methods, they end up generating pairs of short integrands and long integrals or vice versa. For all three data generation methods, they also end up producing integrands that are similar to each other. By similar, we mean expressions that are the same up to the coefficients of an expression. These issues cause a model to overfit on the training data.

\subsection{New Methods}
\label{sec:data/new}

We presented a new data generation method, \textbf{RISCH}, for elementary integrable expressions in \cite{Barket2023_generation}. This is based on the Risch method for symbolic integration and was shown to generate different types of integrals from the previous three.% methods as well as address the shortcomings they suffer from. 

We have developed another method, \textbf{SUB}, which like IBP uses an existing method to generate more data: this time the substitution rule.
\begin{theorem}[Substitution Rule]\label{theorem:sub}
If $u = g(x)$ is a differentiable function whose range is an interval $I$ and $f$ is continuous on $I$, then
    \begin{equation*}
        \int f(g(x))g'(x) dx = \int f(u) du.
    \end{equation*}
\end{theorem}
This means that if $g(x)$ is any differentiable function and $f(x)$ is any integrable function, then we can use Theorem \ref{theorem:sub} to generate an (integrand, integral) pair. Indeed, if $\int f=F$, then by Theorem \ref{theorem:sub}, we have $\int f(g(x))g'(x)dx = F(g(x)) + C$. Like the IBP method, SUB requires a dataset of integrable expressions for $f(x)$. We can use the FWD, BWD, and RISCH methods to generate such a dataset. This method is simple but effective at generating many expressions. 

% Like RISCH, a weakness of SUB is that many of the integrands use a specific sub-algorithm in Maple which leads to bias in the data. We must be careful about how many examples to include from the RISCH and SUB methods to balance the dataset.

\subsection{Datasets of Integrals}

Existing datasets of integrals were thoroughly discussed in Section 2.2 of \cite{Barket2023_generation}.

Maplesoft curates a dataset for indefinite integration for testing their integration function during each release: 47,750 examples exist ranging from very simple expressions to complex expressions with many parameters and special functions. Rather than train on this data, we will keep it aside for validation. This is important to show that ML models can generalise outside of the data they are trained on. 
As our domain for the data generation methods only consists of elementary functions, we only use the 7413 elementary expressions in the Maple test suite for validation at the moment. 

\section{Experimental Results}
\label{sec:Results}

\subsection{Experiment Setup}
\label{sec:results/setup}

\subsubsection{Dataset Preparation:} Before training any model, we must first prepare a dataset to train on. In Section \ref{sec:Data} we discussed five different data generation methods: FWD, BWD, IBP, RISCH, and SUB. The ML models will be trained on 20,000 examples from each of the five methods for a total of 100,000 samples.

\begin{figure}[t]
    \centering
    \includegraphics[scale=0.7]{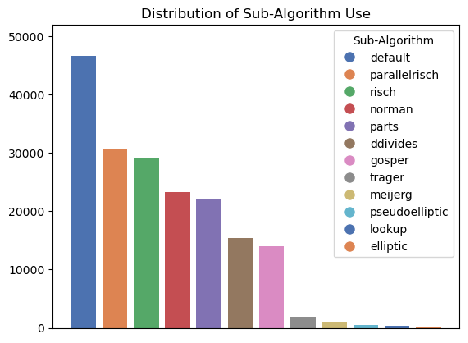}
    \caption{Optimal sub-algorithms for $100$k integrands. %Note that more than one sub-algorithm could be optimal for a given integrand.
    }
    \label{fig:data_freq}
\end{figure}

To label the data, we must first decide on what it means to be optimal.  Maple stores all its expressions as a directed acyclic graph (DAG). It differs from a binary tree representation in that whenever a node is repeated, it does not make a new child for that node. Rather, any common sub-expressions only generate one node in memory and are referenced in multiple places where they occur in the expression. This is useful for avoiding redundant storage of identical sub-expressions. To measure the DAG size, we use a custom Maple function: \verb|expr -> length(sprintf("%lm",expr))|. This is measuring the length of the serialized format of a DAG in Maple. We optimise on this when training ML.  
% Labels do not come for free: to figure out which is the best sub-algorithm for each integrand, we must run each expression through Maple for each of the sub-algorithms and record the size of each output. 
Figure \ref{fig:data_freq} shows the distribution of optimal sub-algorithms.  The dataset is imbalanced:  %We observe a lack of examples for some of the sub-algorithms. %The first seven algorithms have sufficient data but this starts to tail off with the Gosper sub-algorithm and beyond. 
more data generation ideas and data balancing tools are future work.
%The RISCH generator is able to generate many examples for the risch and parallelrisch sub-algorithms, as well as SUB and derivativedivides. More data generators should be created in the future for specific sub-algorithms to balance the dataset.

% Due to this, we are going to examine precision as well as accuracy when evaluating our models. Precision is a good metric to examine in our case because it measures the accuracy of positive predictions (sub-algorithms that are optimal). If there are a lot of false positives, then our precision score will be low. In this setting, a false positive can have detrimental effects compared to false negatives. This is because if we predict a sub-algorithm this is not optimal, there is a high possibility of getting a non-optimal answer. On the other hand, if we miss an algorithm that was actually optimal, we can still get an optimal answer from one of the other sub-algorithms due to this being a multi-label classification problem.

% We also note that we do not train any models for the last three sub-algorithms: Lookup, Elliptic, and Pseudoelliptic. To understand why, we must divert a bit and explain how the Maple integrator works. Before any expression tries one of the sub-algorithms, there is what is called a guard for the sub-algorithm. The guard is responsible for doing a quick check on the expression to see if the sub-algorithm will even be applicable to the given input. On top of having a severe lack of data for these three algorithms, the guards implemented for these are efficient and, if the sub-algorithm is applicable, usually tend to already produce the best answer. Rather than training a model, we will instead just utilise the guards and return their answer if their sub-algorithm is applicable.

\subsubsection{Preprocessing:}
We pre-process the dataset by first removing any expressions where every single algorithm was unsuccessful. %This is because the order we try them will not matter, we will still get a result of failure at the end. 
We then replace the integers within each expression in the following manner:
\begin{enumerate}
    \item if the integer is in the range $[-2, 2]$, then nothing changes;
    \item if the integer is single-digit not in the range $[-2, 2]$, replace by \verb|CONST| token;
    \item if the integer has two digits, replace by a \verb|CONST2| token; and
    \item for all other cases, replace by a \verb|CONST3| token.
\end{enumerate}
The rationale here is that while normally the coefficients of an expression would not change, this is not the case for small exponents ($(x+1)^{1}$ and $(x+1)^{-1}$ integrate very differently). The range $[-2, 2]$ captures these special properties, while for the rest, we only differentiate by the number of digits.  
In \cite{Piotrowski2019} the authors critique \cite{Lample2020} for having many examples equivalent up to constants, risking for data leakage. We only keep unique copies of expressions after replacing the integers with \verb|CONST| tokens, reducing the training data by roughly 10\%.

\subsubsection{Model Design:} As discussed earlier, we perform binary relevance for each sub-algorithm and train both LSTM and TreeLSTM models.  All hyperparameters are kept the same between the two models so we can directly compare the architectures. Note that these hyperparameters have not been fully optimised for these preliminary results.  Each model consists of an embedding layer, and two layers (LSTM or TreeLSTM) with the first having 64 cells and the second having 32 cells. We also include 40\% dropout on the second layer to help avoid overfitting. We then have a fully connected with ReLU activation, and finally a single cell Sigmoid-activated output of the probability. Both LSTM and TreeLSTM were implemented with Pytorch as the backend, and the TreeLSTM utilised a specific graph learning library called Deep Graph Learning \cite{Wang2019_dgl}.

\subsubsection{Evaluation:}   Each expression will have a probability vector given from the model (the probability of the sub-algorithm being optimal). To evaluate our models we let them select the sub-algorithm with the highest probability; and if that algorithm is not successful we try the next most probably. 

We will test against each other and Maple's existing meta-algorithm.  We maintain a separate test set of 25,000 expressions with an equal split from all five generators.  We will also evaluate on the 7413 examples from the Maple test suite to determine if the models can generalise to different data.

\subsection{Experiment Results}

A code implementation of the experiments used to generate the results is available at \url{https://github.com/rbarket/Int\_Algo\_Selection}.

The models took 312s and 178s to train for the TreeLSTM and LSTM models respectively, when averaged over training each binary classifier on a single GPU. 

\subsubsection{Results on the Testing Dataset:}

%\begin{figure}[t]
%\centering
%\begin{subfigure}{.49\textwidth}
%  \centering
%  \includegraphics[width=.95\textwidth]{Figures/total_test.png}
%  \caption{Total Optimal Predictions}
%  \label{fig:counts_test_total}
%\end{subfigure}%
%\begin{subfigure}{.49\textwidth}
%  \centering
%  \includegraphics[width=.95\textwidth]{Figures/unique_test.png}
%  \caption{Unique Optimal Predictions}
%  \label{fig:counts_test_unique}
%\end{subfigure}
%\caption{Comparing the results of the ML predicted answers to Maple's meta-algorithm on 25,000 test cases. Figure (\ref{fig:counts_test_total}) shows the total number of times each one produced the optimal answer or was within a certain margin of producing the optimal answer. Figure (\ref{fig:counts_test_unique}) shows when one of Maple or the models uniquely got the best answer.}
%\label{fig:counts_test}
%\end{figure}

\begin{figure}[t]
\centering
  \includegraphics[width=.7\textwidth]{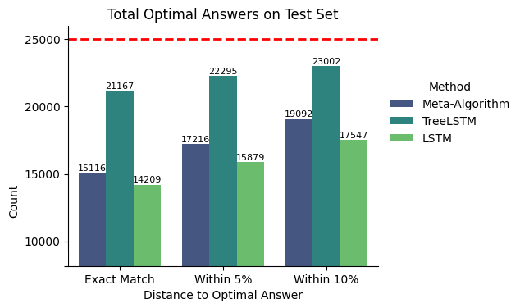}
  \caption{The number of times each ML model and Maple's meta-algorithm produced the optimal answer, or came close to it, on the testing dataset. \label{fig:counts_test_total}}
\end{figure}

Results for the 25,000 test cases are summarised in Figure \ref{fig:counts_test_total}.  
TreeLSTM is the dominant model, predicting 84.6\% of exact optimal answers compared to 60.5\% and 56.8\% from Maple's meta-algorithm and LSTM respectively. This is the case even when we allow for the algorithms to have get only close to the optimal and still be considered a success.

We analysed how many problems each uniquely predict correctly: 441 for Maple's meta-algorithm, 367 for the LSTM and 2631 for the TreeLSTM.  So there are only small quantities of examples where the TreeLSTM is outperformed.

% % Table

% \begin{table}[t]
% \centering
% \begin{tabular}{c|c|c|c}

%  Integrand & Maple & TreeLSTM & LSTM  \\ \hline
%  $\frac{1}{x +x^{x}}-\frac{x \left(1+x^{x} \left(\ln \! \left(x \right)+1\right)\right)}{\left(x +x^{x}\right)^{2}}$ & $-\frac{{\mathrm e}^{x \ln \left(x \right)}}{x +{\mathrm e}^{x \ln \left(x \right)}}$ & $\frac{x}{x +x^{x}}$ & $\frac{x}{x +{\mathrm e}^{x \ln \left(x \right)}}$  \\ \hline
%  &  &  &   \\ \hline
%  &  &  &   \\ 
% \end{tabular}
% \caption{NOTE: Table incomplete, still need to fill this in}
% \end{table}

This results are promising, especially considering the low amount of training time and relatively small size of the dataset (compared to other ML applications). Scaling both these factors up should improve performance further. This clearly validates our hypothesis on the benefits of the tree-based representation for mathematical expressions when compared to a sequence of tokens. Before we conclude the benefits of an ML approach, we should judge generalisability of the models by validating performance on independent data.

\subsubsection{Results on the Maple Test Suite:}

%\begin{figure}[ht]
%\centering
%\begin{subfigure}{.49\textwidth}
%  \centering
%  \includegraphics[width=.95\textwidth]{Figures/total_maple.png}
%  \caption{Total Optimal Predictions}
%  \label{fig:counts_maple_total}
%\end{subfigure}
%\begin{subfigure}{.49\textwidth}
%  \centering
%  \includegraphics[width=.95\textwidth]{Figures/unique_maple.png}
%  \caption{Unique Optimal Predictions}
%  \label{fig:counts_maple_unique}
%\end{subfigure}

\begin{figure}[t]
  \centering
  \includegraphics[width=.7\textwidth]{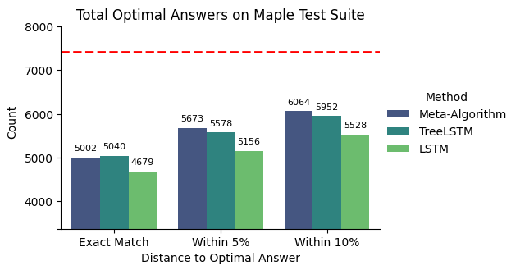}
\caption{The number of times each ML model and Maple’s meta-algorithm produced the optimal answer, or came close to it, on the Maple Test Suite. \label{fig:counts_maple_total}
}
\end{figure}

Performance on the Maple test suite is summarised in Figure \ref{fig:counts_maple_total} and paints a somewhat different picture. The TreeLSTM still has the most optimal answers, however, this time allowing a 5\% or 10\% margin of error lets Maple's meta-algorithm do better.  Both outperform the LSTM.  These results demonstrate that the ML models generalise to independently sourced data. Comparing the Maple's meta-algorithm only to the TreeLSTM:  the former has 578 uniquely optimal answers and the latter 633.  

\section{Conclusion}
\label{sec:Conclusion}

We are able train a ML model to do sub-algorithm selection for symbolic integration to outperform the current state-of-the-art meta-algorithm for the task in Maple. 
The representation of our data plays a crucial role: the TreeLSTM and LSTM models were the same up to their unique architecture layers, demonstrating the benefit of a tree embedding over a simple sequence of tokens. %Moreover, the TreeLSTM produces many more unique optimal answers compared to Maple and LSTM.   

Importantly, the TreeLSTM is also competitive with Maple's meta-algorithm on data produced independently from the training set. This is important to show the value of pursuing such an approach for use by Maple in a general-purpose integration routine.  Such generalisation was something \cite{Lample2020} failed to demonstrate.  

We are confident that with an increase in the quantity of training data, and hyperparameter optimisation, the ML models can improve further still.  

\bibliographystyle{splncs04}
\bibliography{ref.bib}

\end{document}